\documentclass[nohyperref]{article}

\usepackage{microtype}
\usepackage{graphicx}
\usepackage{subfigure}
\usepackage{booktabs} 

\usepackage{hyperref}

\usepackage[accepted]{icml2022}

\usepackage{amsmath}
\usepackage{amssymb}
\usepackage{mathtools}
\usepackage{amsthm}

\usepackage[capitalize,noabbrev]{cleveref}

\theoremstyle{plain}

\theoremstyle{definition}

\theoremstyle{remark}

\usepackage[textsize=tiny]{todonotes}

\usepackage{hyperref}
\usepackage{url}
\usepackage{wrapfig}
\usepackage{multirow}
\usepackage{nicefrac}
\usepackage{booktabs}
\usepackage{makecell}
\usepackage{colortbl}
\newcommand{\matr}[1]{\mathbf{#1}}

\usepackage{enumitem}
\setlist[itemize]{leftmargin=0.5cm}

\usepackage{caption}
\usepackage{xcolor}

\newcommand{\hr}[1]{\textcolor{black}{#1}}

\icmltitlerunning{ShiftAddNAS: Hardware-Inspired Search for More Accurate and Efficient Neural Networks}

\begin{document}

\twocolumn[
\icmltitle{
ShiftAddNAS: Hardware-Inspired Search \\for More Accurate and Efficient Neural Networks
}



\icmlsetsymbol{equal}{*}

\begin{icmlauthorlist}
\icmlauthor{Haoran You$^{\dagger}$}{rice}
\icmlauthor{Baopu Li}{oracle}
\icmlauthor{Huihong Shi}{rice}
\icmlauthor{Yonggan Fu}{rice}
\icmlauthor{Yingyan Lin}{rice}
\end{icmlauthorlist}

\icmlaffiliation{rice}{Rice University}
\icmlaffiliation{oracle}{Oracle Health and AI}

\icmlcorrespondingauthor{Baopu Li}{baopu.li@oracle.com}
\icmlcorrespondingauthor{Yingyan Lin}{yingyan.lin@rice.edu}

\icmlkeywords{Machine Learning, ICML}

\vskip 0.3in
]



\printAffiliationsAndNotice{$^\dagger$Work done while interning at Baidu USA}

\begin{abstract}
Neural networks (NNs) with intensive multiplications (e.g., convolutions and transformers) are capable yet power hungry, impeding their more extensive deployment into resource-constrained devices.
As such, 
multiplication-free networks, which follow a common practice in energy-efficient hardware implementation to parameterize NNs with more efficient operators (e.g., bitwise shifts and additions), have gained growing attention. However, multiplication-free networks usually under-perform their vanilla counterparts in terms of the achieved accuracy. To this end, this work advocates hybrid NNs that consist of both powerful yet costly multiplications and efficient yet less powerful operators for marrying the best of both worlds, and proposes \textbf{ShiftAddNAS}, which can automatically search for more accurate and more efficient NNs. Our ShiftAddNAS highlights two enablers. Specifically, it integrates (1) the first hybrid search space that incorporates both multiplication-based and multiplication-free operators for facilitating the development of both accurate and efficient hybrid NNs; and (2) a novel weight sharing strategy that enables effective weight sharing among different operators that follow heterogeneous distributions (e.g., Gaussian for convolutions vs. Laplacian for add operators) 
and simultaneously leads to a largely reduced supernet size and much better searched networks.
Extensive experiments and ablation studies on various models, datasets, and tasks consistently validate the efficacy of ShiftAddNAS, e.g., 
achieving up to a \textbf{+7.7\%} higher accuracy or a \textbf{+4.9} better BLEU score compared to state-of-the-art NN, 
while leading to up to \textbf{93\%} or \textbf{69\%} energy and latency savings, respectively.
Codes and pretrained models are available at \url{https://github.com/RICE-EIC/ShiftAddNAS}.

\end{abstract}

\vspace{-1em}
\section{Introduction}

The unprecedented performance achieved by neural networks (NNs), e.g., convolutional neural networks (CNNs) and Transformers, requires intensive multiplications and thus prohibitive training and inference costs, contradicting the explosive demand for embedding various intelligent functionalities into pervasive resource-constrained edge devices.
In response, multiplication-free networks have been proposed to alleviate the prohibitive resource requirements by replacing the costly multiplications with lower-cost operators for boosting hardware efficiency. 
For example, 
AdderNet \citep{chen2019addernet} utilizes mere additions to design NNs; and ShiftAddNet \citep{ShiftAddNet} 
follows a commonly used hardware practice to re-parameterize NNs with both bitwise shifts and additions. 
Despite their promising performance in hardware efficiency, multiplication-free NNs in general under-perform their CNN and Transformer counterparts in terms of task accuracy for both computer vision (CV) and natural language processing (NLP) applications.

To marry the best of both worlds, we advocate hybrid multiplication-reduced network architectures that integrate both multiplication-based operators (e.g., vanilla convolution \citep{krizhevsky2012imagenet} and attention \citep{transformer}) and multiplication-free operators (e.g., shift and add \citep{ShiftAddNet}) to simultaneously boost task accuracy and efficiency. Thanks to the amazing success of neural architecture search (NAS) in automating the  process of designing state-of-the-art (SOTA) NNs, it is natural to consider NAS as the design engine of the aforementioned hybrid NNs for various applications and tasks, each often requiring a different performance-efficiency trade-off.
However, there still exist a few challenges in leveraging NAS to design the hybrid NNs.
\textit{\underline{\textit{First}}}, existing NAS methods mostly consider the search for either efficient CNNs \citep{wan2020fbnetv2}, Transformers \citep{AutoFormer}, or hybrid CNN-Transformers \citep{HR-NAS,li2021bossnas}, and there still is  \textit{a lack of a seminal work that searches for multiplication-reduced hybrid networks, especially for the hardware-inspired networks that incorporate both bitwise shifts and additions.}
\textit{\underline{Second}}, a hybrid search space could make it more challenging to achieve effective NAS and further aggravate the search burden, due to the enlarged search space imposed by the newly introduced multiplication-free operators. It is worth noting that \textit{existing weight sharing strategies of NAS cannot directly be applied to the target hybrid search space, because weights of different operators follow heterogeneous distributions, leading to a dilemma of either inefficient search or inconsistent architecture ranking}. Specifically, weights in convolutional and adder layers follow Gaussian and Laplacian distributions, respectively, as also highlighted by \citep{chen2019addernet,adder_distillation}. As such, forcing weight sharing among heterogeneous operators could hurt  
the capacity and thus the achieved accuracy of the resulting NNs, while treating them separately could explode the search space and make it more difficult to achieve effective NAS, i.e., the dilemma mentioned above.   

To tackle the aforementioned challenges towards more accurate and efficient NNs, this work makes the following contributions:

\vspace{-0.3em}
\begin{enumerate}[leftmargin=5mm,topsep=0mm]
\itemsep -0.3\parsep

    \item We propose a generic NAS framework dubbed \textbf{ShiftAddNAS}, which for the first time can automatically search for 
    efficient hybrid NNs with both superior accuracy and efficiency. Our ShiftAddNAS integrates a hybrid hardware-inspired search space that incorporates both multiplication-based operators (e.g., convolution and attention) and multiplication-free operators (e.g., shift and add), and can serve as a play-and-plug module to be applied on top of SOTA NAS works for further boosting their achievable accuracy and efficiency. 
    
    \item We develop a new weight sharing strategy for effective search with hybrid search spaces, which only incurs a negligible overhead when searching for hybrid operators with heterogeneous (e.g., Gaussian vs. Laplacian) weight distributions as compared to the vanilla NAS with merely multiplication-based operators, alleviating the dilemma mentioned above regarding either inefficient search or inconsistent architecture ranking. 
    
    \item We conduct extensive experiments and ablation studies to validate the effectiveness of ShiftAddNAS against SOTA works. Results on multiple benchmarks demonstrate the superior accuracy and hardware efficiency of its searched NNs as compared to both (1) manually designed multiplication-free networks, CNNs, Transformers, and hybrid CNN-Transformers, and (2) SOTA NAS works, on both CV and NLP tasks.
    
\end{enumerate}
\vspace{-0.5em}

\section{Related Works}

\textbf{Multiplication-free NNs.}
Many efficient NNs aim to reduce their intensive multiplications that dominate the time/energy costs.
One important trend is to replace the multiplications with lower-cost operators: 
BNNs \citep{courbariaux2016binarized,juefei2017local} binarize both the weights and activations and reduce multiplications to merely sign flips at non-negligible accuracy drops;
AdderNets \citep{chen2019addernet,adder_distillation,adder_hardware} fully replace the multiplications with lower-cost additions and further develop an effective backpropagation scheme for efficient AdderNet training;
Shift-based NNs leverage either spatial shift \citep{wu2018shift} or bit-wise shift operations, e.g., DeepShift \citep{deepshift}, to reduce the amount of multiplications;
and 
ShiftAddNet \citep{ShiftAddNet} draws inspirations from efficient hardware designs to reparamatize NNs with mere bitwise shifts and additions. There are also dedicated architecture designs for supporting such multiplication-free networks \citep{adder_hardware,inci2022quidam,inci2022qappa,inci2022qadam}.
While multiplication-free NNs under-perform their vanilla NN counterparts in terms of achieved accuracy, ShiftAddNAS aims to automatically search for multiplication-reduced NNs that incorporate both multiplication-based and multiplication-free operators for marrying the best of both worlds, i.e., boosted accuracy and efficiency.

\textbf{Neural architecture search.}
NAS has achieved an amazing success in automating the design of efficient NN architectures. 
For searching for CNNs, early works \citep{tan2019efficientnet,tan2019mnasnet,howard2019searching} adopt reinforcement learning based methods that require a prohibitive search time and computing resources,
while recent works \citep{liu2018darts,wu2019fbnet,wan2020fbnetv2,nas_cvpr2021} utilize differentiable search to greatly improve the search efficiency.
More recently, some works adopt one-shot NAS \citep{guo2020single,cai2019once,yu2020bignas,wang2021alphanet} to decouple the architecture search from supernet training and then evaluate the performance of sub-networks whose weights are directly inherited from the pretrained supernet.
For searching better Transformers, recently emerging works \citep{wang2020hat,VITAS,AutoFormer, GLiTICCV21} take one-shot NAS and an evolutionary algorithm to obtain optimal Transformer architectures for both NLP and CV tasks.
Additionally, BossNAS \citep{li2021bossnas} and HR-NAS \citep{HR-NAS} further search for hybrid CNN-Transformer architectures.

\vspace{-0.2em}
Nevertheless, little effort has been made to explore NAS methods especially their search strategies for multiplication-reduced NNs.
Furthermore, it is not clear whether existing efficient NAS methods are applicable to search for such multiplication-reduced NNs. 
As such, it is highly desirable to develop NAS methods, e.g., ShiftAddNAS, dedicated for hardware-inspired multiplication-reduced NNs to increase achievable accuracy and efficiency.

\textbf{Transformers.}
Transformers \citep{transformer} were first proposed for NLP tasks, which have inspired many interesting works.
Some advance Transformer architecture by improving the attention mechanism \citep{transformer_improve}, training deeper Transformers \citep{transformer_deep}, and replacing the attention with convolutional modules \citep{wu2018pay}; and others strive to reduce Transformers' computational complexity by adopting sparse attention mechanisms \citep{bigbird}, low-rank approximation \citep{wang2020linformer}, or compression techniques \citep{lite_transformer}.
Recently, there has been a growing interest in developing Transformers for CV tasks. Vision Transformer (ViT) \citep{vit} for the first time successfully applies pure Transformers to image classification and achieves SOTA task accuracy, which yet relies on pretraining on giant datasets. 
Following works including DeiT \citep{DeiT}, the authors in T2T-ViT \citep{yuan2021tokens} develop new training recipes and tokenization schemes, for achieving comparable accuracy without the necessity of costly pretraining;
and another trend is to incorporate CNN modules into Transformer architectures for better accuracy and efficiency tradeoffs \citep{wu2021cvt,graham2021levit}.
In contrast, we advocate hybrid multiplication-reduced NNs and develop an automated search framework that can automatically search for such hardware inspired hybrid models.

\vspace{-0.5em}
\section{The Proposed ShiftAddNAS Framework}
\vspace{-0.2em}

In this section, 
we first introduce the hybrid search space from both algorithmic and hardware cost perspectives, providing high-level background and justification for motivating ShiftAddNAS; Sec.~\ref{sec:search_method} elaborates the one-shot search method of ShiftAddNAS by first analyzing the dilemma of either inefficient search or inconsistent architecture ranking and then introducing the proposed novel heterogeneous weight sharing strategy tackling the aforementioned dilemma.

\vspace{-0.3em}
\subsection{ShiftAddNAS: Hybrid Search Space}
\label{sec:search_space}
\vspace{-0.2em}

\begin{figure*}[t]
    \centering
    \includegraphics[width=\textwidth]{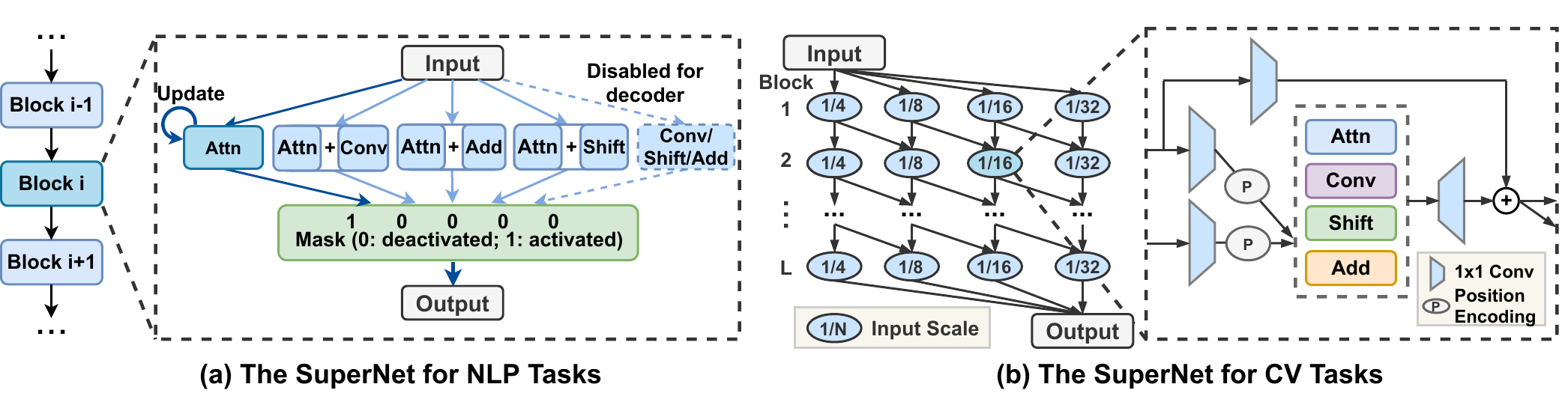}
    \vspace{-1.7em}
    \caption{Supernets for NLP and CV tasks: (a) For NLP, we adopt a multi-branch structure for each block of the supernet, where Attn+Conv represents the channel-wise concatenation of these two blocks, and (b) for CV tasks, we consider a multi-resolution pipeline for each block of the supernet. }
    \label{fig:supernets}
    \vspace{-1em}
\end{figure*}

\textbf{Candidate blocks.}
The first step of developing ShiftAddNAS is to construct a hybrid search space incorporating suitable building blocks that exhibit various performance-efficiency trade-offs.
Specifically, we hypothesize that integrating both multiplication-based and multiplication-free blocks into the search space could lead to both boosted accuracy and efficiency, and consider blocks from two trends of designing NNs: 
(1) \textit{capable} NNs, e.g., vanilla CNNs and Transformers, leverage either convolutions ({\tt \textbf{Conv}}) or multi-head self-attentions ({\tt \textbf{Attn}}) that comprise of intensive multiplications to capture 
local or 
global correlations, achieving a SOTA accuracy in both CV and NLP tasks; and 
(2) \textit{efficient} multiplication-free NNs, e.g, ShiftAddNet, draw inspirations from hardware design practices to incorporate two 
efficient and complementary blocks, i.e., coarse-grained {\tt \textbf{Shift}} and fine-grained {\tt \textbf{Add}}, for 
favoring hardware efficiency, while maintaining a decent accuracy. 
While our constructed general hybrid search space for both NLP and CV tasks are shown in Fig.~\ref{fig:supernets}, we next analyze the building blocks from both algorithmic and hardware costs perspectives.

{\tt \textbf{Attn}} is a core component of Transformers \citep{transformer}, which consists of a number of heads $\text{H}$ with each capturing different global-context information by measuring pairwise correlations among tokens as defined below:
\vspace{-0.3em}
\begin{equation} \label{equ:attn_op}
\begin{split}
\textit{\textbf{O}}_{\tt \textbf{Attn}} \!=\! {\tt Concat}(\text{H}_1, \cdots, \text{H}_h) \cdot W^O,
\,  \mathrm{where} \,\, \\
\text{H}_i \!=\! {\tt Softmax}(\frac{QW_i^Q \cdot (KW_i^K)^T}{\sqrt{d_k}}) \cdot VW_i^V,
\end{split}
\vspace{-1em}
\end{equation}
where $h$ denotes the number of heads, $Q, K, V \in \mathbb{R}^{n \times d}$ are the query, key, and value embeddings of hidden dimension $d$ obtained by linearly projecting the input sequence of length $n$.
In this way, the {\tt \textbf{Attn}} block first computes dot-products between key-query pairs, scales to stabilize the training, uses ${\tt Softmax}$ to normalize the resulting attention scores, and then computes a weighted sum of the value embeddings corresponding to different inputs.
Finally, the results from all heads are concatenated and further projected with a weight matrix $W^O \in \mathbb{R}^{d \times d}$ to generate the outputs.

{\tt \textbf{Conv}} is a key operator of CNNs, which models the local-context information of high-dimensional inputs such as images through sliding kernel weights $W$ on top of inputs $X$ to measure their similarity \citep{gu2018recent}, as defined in Eq. (\ref{equ:conv_op}).
    Its  
    translation invariant and weight sharing ability leads to various SOTA CNNs \citep{he2016deep} or hybrid CNN-Transformer models \citep{xiao2021early}. However,
    the computational complexities of CNNs can be prohibitive due to their  intensive multiplications. For example, one forward pass of ResNet-50 \citep{he2016deep} requires 4G floating point multiplications.
    \vspace{-0.5em}
    \begin{equation} \label{equ:conv_op}
    \begin{split}
    \textit{\textbf{O}}_{\tt \textbf{Conv}} &= \sum X^T * W
    \end{split}
    \end{equation}
    \vspace{-2em}

{\tt \textbf{Shift}} is a well-known efficient hardware primitive, motivating the recent development of shift-based efficient NNs. For example, DeepShift \citep{deepshift} parametrizes NNs with bitwise shifts and sign flips, as formulated in Eq. (\ref{equ:shift_op}), with $W = S \cdot 2^P$ denoting weights in the shift blocks, where $S \in \{-1, 0, 1\}$ are sign flip operators and the power-of-two parameter for $P$ represents the bitwise shifts. However, NNs built with shift blocks and quantized weights are observed to be inferior to multiplication-based NNs in terms of expressiveness (accuracy) as validated in \citep{ShiftAddNet}.
    \vspace{-0.5em}
    \begin{equation} \label{equ:shift_op}
    \begin{split}
    \textit{\textbf{O}}_{\tt \textbf{Shift}} = \sum X^T * (S \cdot 2^P) 
    \end{split}
    \end{equation}
    \vspace{-2em}
    
\begin{figure}[t]
\resizebox{\linewidth}{!}{
    \includegraphics[width=\linewidth]{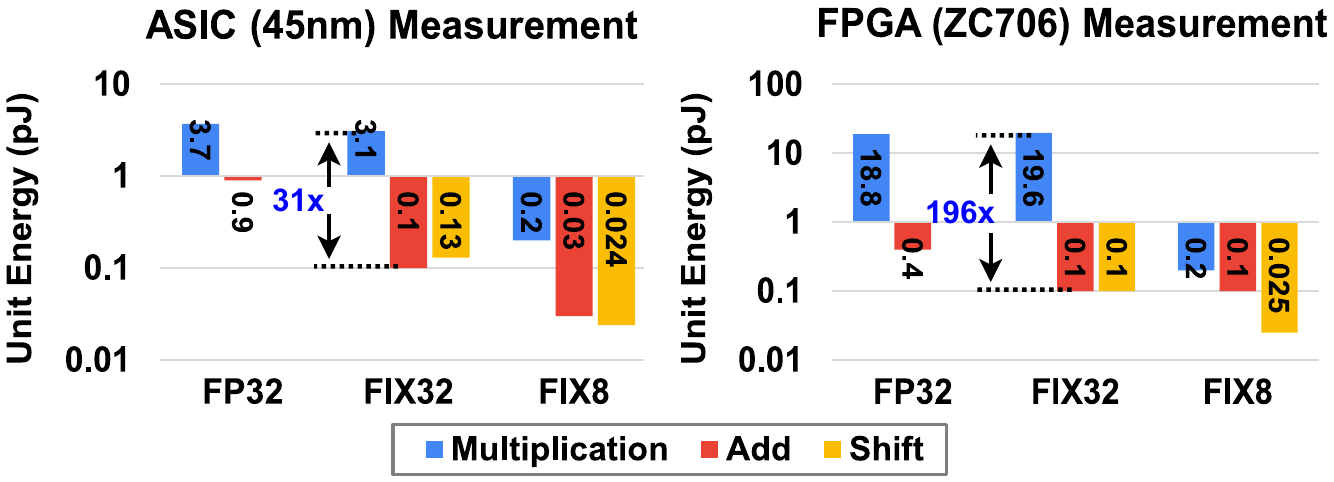}
}
\vspace{-2em}
\caption{Unit energy comparisons.}
\vspace{-0.5em}
\label{fig:unit_energy}
\end{figure}

\begin{table}[t]
\centering
\vspace{-0em}
      \caption{The search space for NLP tasks.}
      \resizebox{\linewidth}{!}{
        \begin{tabular}{c||c}
        \hline
        \multirow{2}{*}{Encoder block types} & [Attn, Attn+Conv, Attn+Shift] \\ & [Attn+Add, Conv, Shift, Add] \\
        \hline
        \multirow{2}{*}{Decoder block types} & [Attn, Attn+Conv] \\ & [Attn+Shift, Attn+Add] \\
        \hline
        Num. of decoder blocks & [6, 5, 4, 3, 2, 1] \\
        \hline
        Elastic embed. dim. & [1024, 768, 512] \\
        \hline
        Elastic head num. & [16, 8, 4] \\
        \hline
        Elastic MLP dim. & [4096, 3072, 2048, 1024] \\
        \hline
        Arbitrary 
        Attn & [3, 2, 1] \\
        \hline
        \end{tabular}%
        \label{tab:NLP_space}%
        }
    \vspace{-1.2em}
\end{table}

{\tt \textbf{Add}} is another efficient hardware primitive which motivates recent 
works \citep{chen2019addernet, adder_hardware, song2021addersr} to design efficient NNs using merely additions to measure the similarity between kernel weights $W$ and inputs $X$, as shown in Eq. (\ref{equ:add_op}). Such add-based NNs \citep{chen2019addernet,adder_distillation} in general have a better expressive capacity than their shift-based counterparts. For example, AdderNets \citep{chen2019addernet} achieve a 1.37\% higher accuracy than DeepShift under similar or even lower FLOPs on ResNet-18 with the ImageNet dataset. However, add-based operators (i.e., repeated additions) are not parameter-efficient as compared to bitwise shift operations \citep{ShiftAddNet}. 
While NNs combining shfit and add achieve a boosted accuracy, efficiency, and robustness than NNs using merely either of them, their accuracy still compares unfavorably in contrast to vanilla CNNs or Transformers.
\vspace{-0.5em}
\begin{equation} \label{equ:add_op}
\begin{split}
\textit{\textbf{O}}_{\tt \textbf{Add}} = - \sum \| X - W \|_1 
\end{split}
\end{equation}
\vspace{-2em}

Based on the above introduction, the search space in ShiftAddNAS incorporates all the four different types of blocks (i.e., Attn, Conv, Shift, and Add), aiming to push forward both NNs' accuracy and efficiency. Note that we refer to all operators as blocks, and adopt block based search space because it has been evidenced and proven that block based ones can reduce the search space size and lead to more accurate architecture ranking/rating ~\citep{li2019improving, li2020blockwisely}.

\textbf{Hardware cost.} As mentioned, 
multiplication-based operators (e.g., {\tt \textbf{Attn}} and {\tt \textbf{Conv}}) favor a superior accuracy yet is not hardware efficient, while multiplication-free operators (e.g., {\tt \textbf{Shift}} and {\tt \textbf{Add}}) favor a better hardware efficiency yet can hurt the achievable accuracy. Specifically, as shown in Fig. \ref{fig:unit_energy}, bitwise shifts can save as high as 196$\times$ and 24$\times$ energy costs over multiplications, when implemented in a 45nm CMOS technology and SOTA FPGA \citep{zc706}, respectively; with a 16-bit precision, bitwise shifts may achieve at least 9.7$\times$ and 1.45$\times$ average power and area savings than multipliers \citep{deepshift}; and 
similarly, additions can save up to 196$\times$ and 31$\times$ energy costs over multiplications in 32-bit fixed-point (FIX32) formats, and 47$\times$ and 4.1$\times$ energy costs in 32-bit floating-point (FP32) formats, when implemented in a 45nm CMOS technology and SOTA FPGA \citep{zc706}, respectively, while 
leading to 1.84$\times$, 25.5$\times$, and 7.83$\times$ area savings than multiplications in a 45nm CMOS technology with FP32, FIX32, and FIX8 formats, respectively \citep{chen2021aqd}.

\textbf{Supernet for NLP tasks.}
Based on the above search space, we construct a supernet for the convenience of search following SOTA one-shot NAS methods ~\citep{cai2018proxylessnas,guo2020single} by estimating the performance of each candidate hybrid model (i.e., subnet) without fully training it. 
As shown in Fig. \ref{fig:supernets} (a), each  
macro-block in the supernet includes all the aforementioned four candidate blocks and 
three multi-branch combinations (e.g., Attn+Conv) along the channel dimension for capturing both global and local context information following \citep{lite_transformer},
where the candidate blocks in the same layer are isolated with each followed by two-layer MLPs and enable elastic embedding dimension, head numbers, and MLP hidden dimension for fine-grained search for efficient NNs as \citep{wang2020hat}. 
Overall, our supernet for NLP tasks contains about $10^{14}$ subnet candidates,
and the searchable choices are listed in Tab. \ref{tab:NLP_space}. During training, all possible subnets are uniformly sampled and only one path is activated for each layer at run-time considering the practical concern on memory consumption for supernet training.
For ease of evaluation, we incorporate common treatments of NAS in our suppenet design. First, for the elastic dimensions mentioned above, all subnets share the front portion of weight channels or attention heads of the largest dimension. Second,
all decoder blocks can take the last one, two, or three encoder blocks as inputs for abstracting both high-level and low-level information \citep{wang2020hat}. 
Note that the number of decoder blocks are also searchable and the conv, shift and add operators are disabled for decoder blocks, as they are observed to be sensitive and activating those paths might hurt the accuracy \citep{ShiftAddNet,wu2018pay}.

\begin{table}[t]
\centering
      \caption{The search space for CV tasks.}
      \resizebox{\linewidth}{!}{
        \begin{tabular}{c||c}
        \hline
        Block types & [Attn, Conv, Shift, Add] \\
        \hline
        Num. of $56^2\times128$ blocks & [1, 2, 3, 4] \\
        \hline
        Num. of $28^2\times256$ blocks & [1, 2, 3, 4] \\
        \hline
        Num. of $14^2\times512$ blocks & [3, 4, 5, 6, 7] \\
        \hline
        Num. of $7^2\times1024$ blocks & [4, 5, 6, 7, 8, 9] \\
        \hline
        \end{tabular}%
      }
      \label{tab:CV_space}
      \vspace{-1.5em}
\end{table}

\textbf{Supernet for CV tasks.}
Different from the commonly used elastic hidden dimension design for NLP tasks, various spatial resolutions or scales are essential for CV tasks.
As such, to ensure more capable feature description of the searched NNs,
we adopt a multi-resolution supernet design. As shown in Fig. \ref{fig:supernets} (b), the supernet incorporates flexible downsampling options, where the spatial resolution for each layer can either stay unchanged or be reduced to half of its previous layer's scale until reaching the smallest resolution. In this way, the four candidate blocks can work collaboratively to deliver the multiscale features required by most CV tasks. Overall, our supernet contains about $10^9$ subnets, for which the detailed searchable choices are summarized in Tab. \ref{tab:CV_space}.
Note that the Attn block is followed by two-layer MLPs and we also include a residual connection for each block as inspired by \citep{srinivas2021bottleneck}.
During training, the supernet performs uniform sampling and only activates one path of the chosen resolution and block type for each layer as for the NLP tasks.

\subsection{ShiftAddNAS: Search Method}
\label{sec:search_method}

\subsubsection{Background and Formulation of One-Shot NAS}

We adopt one-shot NAS for improved search efficiency, i.e., assuming that the subnet candidates can directly inherit their weights from the supernet, following SOTA NAS works. Such a strategy is commonly referred as \textit{weight sharing.} Specifically, 
the supernet $\mathcal{N}$ with parameters $\mathbf{W}$ is trained to obtain the weights for all subnets within the search space $\mathcal{S}$.  Since the supernet training and architecture search are decoupled in one-shot NAS, it usually requires two-level optimization: supernet training and architecture evaluation as defined below:
%
\vspace{-0.5em}
\begin{align}
&\mathbf{W}_\mathcal{S} = \underset{\matr{W}}{\arg\min} \ L_{train}(\mathcal{N} (\mathcal S,\matr{W}) ), \label{eq:supernet_train} 
\end{align}
\vspace{-2em}
\begin{align}
&a^* = \underset{a\in {\mathcal S}}{\arg\max}\ ACC_{ val}(\mathcal{N}(a,\mathbf{W}_{\matr {\mathcal S}}(a)) ).
\label{eq:supernet_optimise}
\end{align}
%
where $\mathcal{N}(\mathcal{S},\mathbf{W})$ represents all possible candidate subnets within the search space.
We first train the supernet by uniformly sampling different subnets $a$ from $S$ as formulated in Eq. (\ref{eq:supernet_train}), after which all subnet candidates $a$ directly inherit their corresponding weights $\mathbf{W}_{\mathcal{S}}(a)$ from the supernet $\mathbf{W}_{\mathcal{S}}$. Finally, we evaluate the accuracy $ACC_{val}(.)$ of each path on the validation set and search for the best subnet with the highest accuracy as formulated in Eq. (\ref{eq:supernet_optimise}). 

\subsubsection{Proposed Heterogeneous Weight Sharing Strategy}

\begin{figure}
\centering
\resizebox{0.95\linewidth}{!}{
    \includegraphics[width=\linewidth]{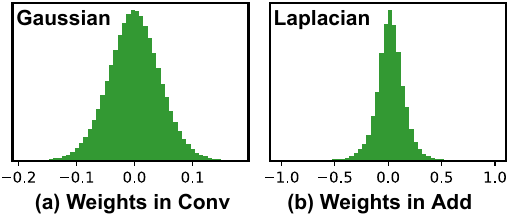}
}
\vspace{-0.5em}
\caption{ Heterogeneous weight distributions in supernets.}
\vspace{-1.8em}
\label{fig:distribution}
\end{figure}


\textbf{Dilemma of vanilla ShiftAddNAS.}
The target hybrid search space of ShiftAddNAS inevitably enlarges the supernet due to the newly considered operators. As such,  activating all block choices without weight sharing as \citep{gong2019mixed,cai2018proxylessnas} can easily explode the memory consumption of NAS. On the other hand, directly sharing weights among different operators as \citep{AutoFormer} will lead to biased search, especially for our hardware-inspired hybrid search space where weights and activations of different operators follow heterogeneous distributions, e.g., weights of the Conv and Add blocks follow a Gaussian and Laplacian distribution, respectively, as shown in Fig. \ref{fig:distribution} and also highlighted in \citep{chen2019addernet} {(more visualization can be found at Appendix \ref{sec:dist_visualize})}. 
Specifically,
if we follow the existing weight sharing strategy to enforce a homogeneous weight distribution among different operators during training the supernet, the resulting weights will not match the heterogeneous weight distributions of independently trained optimal hybrid subnets. That is to say, for NAS with the target hybrid search space, there exists an optimization gap between the goals of weight sharing optimization and individual subnet optimization, where the former is approximated while the latter is accurate \citep{xie2020weight}.
Hence, naively adopting the homogeneous weight sharing strategy can lead to inconsistent architecture ranking, which is a major issue associated with one-shot NAS as pointed out by ~\citep{chu2019fairnas,you2020greedynas}. 



\begin{figure}[t]
    \centering
    \includegraphics[width=\linewidth]{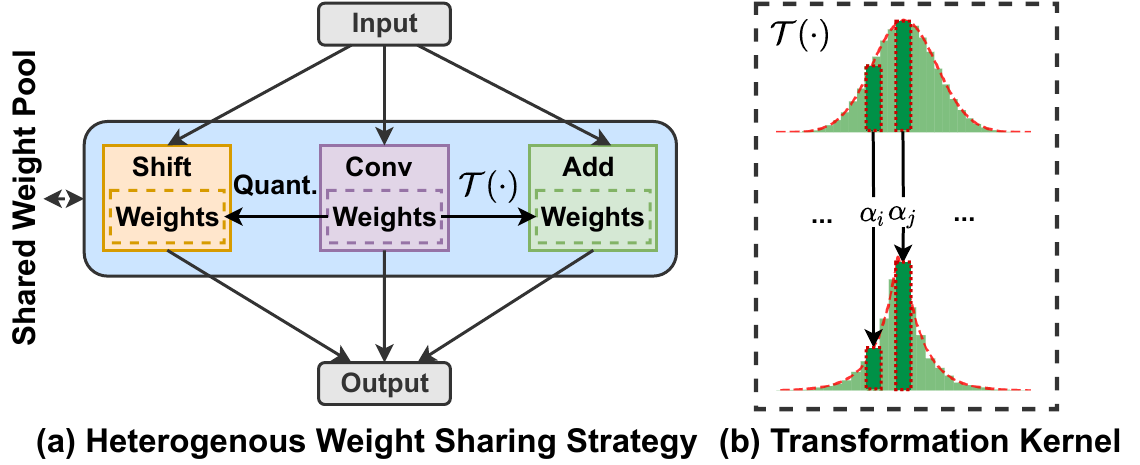}
    \vspace{-1.3em}
    \caption{(a) Illustration of the proposed heterogeneous weight sharing strategy, where weights of shift blocks are quantized to powers of two;  (b)  visualization of the adopted learnable transformation kernel $\mathcal{T}(\cdot)$ for mapping the shared weights of Gaussian distribution to a Laplacian distribution.
    }
    \label{fig:our_weight_sharing}
    \vspace{-1.8em}
\end{figure}

\textbf{Proposed solution: heterogeneous weight sharing.}
To tackle 
the aforementioned dilemma, we propose a heterogeneous weight sharing strategy that can simultaneously reduce the supernet size corresponding to the target hybrid search space and allow weights of different blocks to follow heterogeneous distributions.
Specifically, the learning objective for the supernet includes both the traditional cross-entropy loss and a KL-divergence loss that is to regularize weight distributions to be close to either a standard Gaussian distribution  $\mathcal{N}(0, I)$ or Laplacian distribution $\mathcal{L}_p(0, \lambda)$ \citep{xie2020weight,chen2019addernet}( $I$ is the identity matrix and $\lambda=1$), dedicated for the Conv and Add blocks, respectively, for reducing the aforementioned optimization gap as formulated in Eq. (\ref{equ:objective}):
%
\begin{equation}\label{equ:objective}
\setlength{\abovedisplayskip}{0pt}
\setlength{\belowdisplayskip}{3pt}
    \begin{split}
        & \mathcal{L_S} = \mathcal{L}_{CE} + \mathcal{L}_{KL} = -\frac{1}{N} \sum_{i=1}^{N} P(y_i | x_i) \log(P(\hat{y}_i | x_i)) \\
        & \!\! + \mathcal{D}(P_{\tt Conv}(\mathbf{W}_\mathcal{S}) || \mathcal{N}(0, I)) \!+\! \mathcal{D}(P_{\tt Add}(\mathcal{T}(\mathbf{W}_{\mathcal{S}})) || \mathcal{L}_p(0, \lambda)),
    \end{split}
\end{equation}
where $\{(x_i, y_i)\}_{i=1}^{N}$ are training data, $\hat{y}$ denotes the output prediction, and $\mathcal{D}_{KL}(p||q) \!=\! -\int  p(z) \frac{p(z)}{q(z)} dz$ measures the KL-divergence between two distributions.
During training, 
 we maintain a shared weight pool for each layer to share weights across all the
 {\tt{\textbf{Conv}}}, {\tt{\textbf{Add}}}, and  {\tt{\textbf{Shift}}} blocks, as illustrated in Fig. \ref{fig:our_weight_sharing} (a). Meanwhile,
weights of the {\tt{\textbf{Conv}}} blocks directly leverage the corresponding ones in the shared weight pool for both forward and backpropagation, while being encouraged to follow a Gaussian distribution by the objective function; weights of the
{\tt{\textbf{Shift}}} blocks 
quantize the shared weights of Gaussian distribution to powers of two before multiplying with the input features, we follow \citep{deepshift} to backpropogate the gradients; 
and for the {\tt{\textbf{Add}}} blocks,
we make use of \textbf{a learnable transformation kernel} $\mathcal{T}(\cdot)$ to map the shared weights of Gaussian distribution to a Laplacian distribution. For the learnable transformation kernel as captured by Eq. (\ref{equ:kernel}), 
the core idea is to apply a piece-wise linear transformation after flattening and sorting the weights in a descending order, and then to reshape and rearrange the transformed weights back to their positions before sorting.
%
\begin{equation}\label{equ:kernel}
\setlength{\abovedisplayskip}{0pt}
\setlength{\belowdisplayskip}{3pt}
    \begin{split}
        \mathcal{T}(W) & = \sum_{i=0}^{d-1} \alpha_i \times W_{[s \times i:s \times (i+1)]}, 
    \vspace{-1em}
    \end{split}
\end{equation}
where $\{\alpha_i\}_{i=1}^{d}$ denote the learnable parameters in $\mathcal{T}(\cdot)$, $\{W_i\}_{i=1}^{n}$ represent the sorted weights (a total of $n$) in the pool, $s = \nicefrac{n}{d}$ denotes an interval within which the transformation is linear, as illustrated in Fig. \ref{fig:our_weight_sharing} (b). As validated in our visualization of supernet weights (e.g., Fig. \ref{fig:distribution}), such a transformation kernel can successfully transform the shared weights of Gaussian to the desired Laplacian distribution, which is consistent with previous observations about kernel learning via linear transformation \citep{jain2012metric}. In our design,
each layer has its own learnable kernel $\mathcal{T}(\cdot)$ with a dimension $d$ of 200 throughout all the experiments as we observed that such a dimension is adequate to learn the transformation across all the models and datasets, 
leading to over 40\% supernet size reduction
while only incurring a negligible ($<0.01\%$ of the supernet size and computational costs) search overhead.
After the supernet is well trained, evolution search is applied to find the optimal subnets.



\vspace{-0.5em}
\section{Experiment Results}
\vspace{-0.3em}

In this section, we first describe our experiment setups, and then benchmark ShiftAddNAS over SOTA CNNs, Transformers, and previous NAS frameworks on both NLP and CV tasks. After that, we conduct ablation studies regarding ShiftAddNAS's heterogeneous weight sharing strategy.

\vspace{-0.5em}
\subsection{Experiment Setups}
\vspace{-0.3em}

\begin{table*}[t]
  \centering
\caption{ShiftAddNAS vs. SOTA baselines in terms of accuracy and efficiency on NLP tasks.}
  \renewcommand{\arraystretch}{1.1}
  \resizebox{\textwidth}{!}{
    \begin{tabular}{lccccc||ccccc}
    \Xhline{3\arrayrulewidth}
      & \multicolumn{5}{c||}{WMT'14 En-Fr} & \multicolumn{5}{c}{WMT'14 En-De} \\
\cline{2-11}      & Params & FLOPs & BLEU & Latency & Energy & Params & FLOPs & BLEU & Latency & Energy \\
    \Xhline{3\arrayrulewidth}
    Transformer & 176M & 10.6G & 41.2 & 130ms & 214mJ & 176M & 10.6G & 28.4 & 130ms & 214mJ \\
    Evolved Trans. & 175M & 10.8G & 41.3 & - & - & 47M & 2.9G & 28.2 & - & - \\
    HAT & 48M & 3.4G & 41.4 & 49ms & 81mJ & 44M & 2.7G & 28.2 & 42ms & 69mJ \\
    \textbf{ShiftAddNAS} & \textbf{46M} & \textbf{3.0G} & \textbf{41.8} & \textbf{43ms} & \textbf{71mJ} & \textbf{43M} & \textbf{2.7G} & \textbf{28.2} & \textbf{40ms}  &  \textbf{66mJ} \\
    \hline
    HAT & 46M & 2.9G & 41.1 & 42ms & 69mJ & 36M & 2.2G & 27.6 & 34ms & 56mJ \\
    \textbf{ShiftAddNAS} & \textbf{41M} & \textbf{2.7G} & \textbf{41.6} & \textbf{39ms} & \textbf{64mJ} & \textbf{33M} & \textbf{2.1G} & \textbf{27.8} & \textbf{31ms} & \textbf{52mJ}  \\
    \hline
    HAT & 30M & 1.8G & 39.1 & 29ms & 48mJ & 25M & \textbf{1.5G} & 25.8 & 24ms & 40mJ \\
    \textbf{ShiftAddNAS} & \textbf{29M} & \textbf{1.8G} & \textbf{40.2} & \textbf{16ms} & \textbf{45mJ} & \textbf{25M} & 1.6G & \textbf{26.7} & \textbf{24ms} & \textbf{40mJ} \\
    \hline
    Lite Trans. (8-bit) & 17M & 1G & 39.6 & 19ms & 31mJ & 17M & 1G & 26.5 & 19ms & 31mJ \\
    \textbf{ShiftAddNAS (8-bit)} & \textbf{11M} & \textbf{0.2G} & \textbf{41.5}  & \textbf{11ms} & \textbf{16mJ} & \textbf{17M} & \textbf{0.3G} & \textbf{28.3} & \textbf{16ms} & \textbf{24mJ} \\
    \hline
    Lite Trans. (8-bit) & 12M & 0.7G & 39.1 & 14ms & 24mJ & 12M & 0.7G & 25.6 & 14ms & 24mJ \\
    \textbf{ShiftAddNAS (8-bit)} & \textbf{10M} & \textbf{0.2G} & \textbf{41.1}  & \textbf{10ms} & \textbf{15mJ} & \textbf{12M} & \textbf{0.2G} & \textbf{26.8} & \textbf{9.2ms} & \textbf{14mJ} \\
    \Xhline{3\arrayrulewidth}
    \end{tabular}%
    }
  \label{tab:comp_NLP}%
  \vspace{-1em}
\end{table*}%

\textbf{Datasets, baselines, and evaluation metrics.}
\underline{\textit{For NLP tasks,}} 
we consider two machine translation datasets, WMT'14 English to French (En-Fr) and English to German (En-De), which consist of 36.3M and 4.5M pairs of training sentences, respectively. The train/val/test splits follow the tradition as in \citep{wang2020hat,S2S_learning}.
We consider five baselines: Transformer \citep{transformer}, Lightweight Conv \citep{wu2018pay}, Lite Transformer \citep{lite_transformer}, and two previous NAS works including Evolved Transformer \citep{so2019evolved} and HAT \citep{wang2020hat}.
We evaluate in terms of five evaluation metrics: the number of parameters/FLOPs, achieved BLEU, and hardware energy and latency measured on a SOTA accelerator Eyeriss \citep{chen2016eyeriss} clocked at 250MHz, where the BLEU is calculated with case-sensitive tokenization following \citep{wang2020hat}.
\underline{\textit{For CV tasks,}} 
we consider the ImageNet dataset and four kinds of SOTA baselines: four multiplication-free CNNs \citep{chen2019addernet, adder_distillation, courbariaux2016binarized, deepshift}, two CNNs \citep{he2016deep,hu2018squeeze}, five Transformers \citep{vit,DeiT,yuan2021tokens,TNT,srinivas2021bottleneck}, and four NAS works (i.e., HR-NAS \citep{HR-NAS}, BossNAS~\citep{li2021bossnas}, AutoFormer~\citep{AutoFormer}, and
VITAS~\citep{VITAS}). Similar to those for the NLP tasks, we adopt five evaluation metrics: the number of parameters/MACs, achieved accuracy, and hardware energy and latency. 

\textbf{Search and training settings.}
\underline{\textit{For NLP tasks,}} 
after training the supernet for 40K steps, we adopt an evolutionary algorithm \citep{wang2020hat} to search for subnets with various latency and FLOPs constraints.
During search, we follow \citep{wang2020hat} to adopt a three-layer NN to measure the latency,
which is accurate with an average prediction error of $<$ 5\%.
The searched subnets are then retrained from scratch for another 40K steps.
\underline{\textit{For CV tasks,}}
we follow \citep{AutoFormer} to conduct an evolutionary search with FLOPs constraints for 20 steps.
We train both the supernet and searched subnets using the same recipe and hyperparameters as DeiT \citep{DeiT}.
{More details regarding search and training settings can be found in Appendix \ref{sec:exp_setting}.}

\begin{figure}
    \centering
    \includegraphics[width=\linewidth]{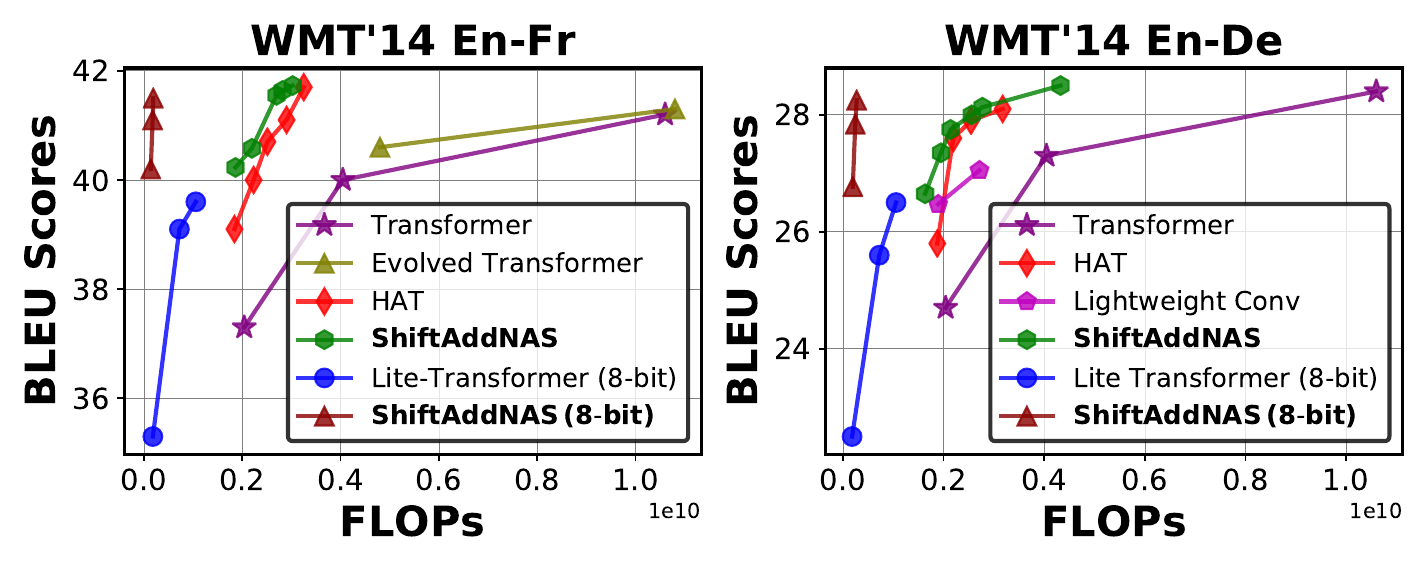}
    \vspace{-1.9em}
    \caption{BLEU scores vs. FLOPs of ShiftAddNAS over SOTA baselines on NLP tasks.}
    \label{fig:comp_NLP}
    \vspace{-2em}
\end{figure}

\begin{table*}[t]
  \centering
  \caption{Comparison with SOTA baselines on ImageNet classification task {(see Tab. \ref{tab:comp_CV_complete} for complete comparisons)}.}
  \renewcommand{\arraystretch}{1.1}
  \resizebox{0.98\textwidth}{!}{
    \begin{tabular}{lcc||ccc||ccc||c}
    \Xhline{3\arrayrulewidth}
    \textbf{Model} & \textbf{Top-1 Acc.} & \textbf{Top-5 Acc.} & \textbf{Params} & \textbf{Res.}  & \textbf{MACs} & \textbf{\#Mult.} & \textbf{\#Add} & \textbf{\#Shift} & \textbf{Model Type} \\
    \Xhline{3\arrayrulewidth}
    AdderNet & 74.9\% & 91.7\% & 26M & $224^2$ & 3.9G & 0.1G & 7.6G & 0 & Mult.-free \\
    DeepShift-PS & 71.9\% & 90.2\% & 52M & $224^2$ & 3.9G & 0.1G & 3.9G & 3.8G & Mult.-free  \\
    ShiftAddNet & 72.3\% & - & 64M & $224^2$ & 10G & 0.1G & 16G & 3.9G & Mult.-free  \\
    \hline
    ResNet-50 & 76.1\% & 92.9\% & 26M & $224^2$ & 3.9G & 3.9G & 3.9G & 0 & CNN  \\
    ResNet-101 & 77.4\% & 94.2\% & 45M & $224^2$ & 7.6G & 7.6G & 7.6G & 0 & CNN \\
    \hline
    ViT-B/16 & 77.9\% & - & 86M & $384^2$ & 18G & 18G & 17G & 0 & Transformer  \\
    DeiT-S & 81.2\% & - & 22M & $224^2$ & 4.6G & 4.6G & 4.6G & 0 & Transformer  \\
    VITAS & 77.4\% & 93.8\% & 13M & $224^2$ & 2.7G & 2.7G & 2.7G & 0 & Transformer  \\
    Autoformer-S & 81.7\% & 95.7\% & 23M & $224^2$ & 5.1G & 5.1G & 5.1G & 0 & Transformer  \\
    \hline
    BoT-50 & 78.3\% & 94.2\% & 21M & $224^2$ & 4.0G & 4.0G & 4.0G & 0 & CNN + Trans.  \\
    BossNAS-T0 & 80.5\% & 95.0\% & 38M & $224^2$ & 3.5G & 3.5G & 3.5G & 0 & CNN + Trans. \\
    \hline
    \textbf{ShiftAddNAS-T0} & \textbf{82.1\%} & \textbf{95.8\%} & 30M & $224^2$ & 3.7G & 2.7G & 3.8G & 1.0G & Hybrid  \\
    \textbf{ShiftAddNAS-T0$\boldsymbol{\uparrow}$} & \textbf{82.6\%} & \textbf{96.2\%} & 30M & $256^2$ & 4.9G & 3.6G & 4.9G & 1.4G & Hybrid  \\
    \Xhline{3\arrayrulewidth}
    T2T-ViT-19 & 81.9\% & - & 39M & $224^2$ & 8.9G & 8.9G & 8.9G & 0 & Transformer  \\
    Autoformer-B & 82.4\% & 95.7\% & 54M & $224^2$ & 11G & 11G & 11G & 0 & Transformer  \\
    \hline
    BoTNet-S1-59 & 81.7\% & 95.8\% & 28M & $224^2$ & 7.3G & 7.3G & 7.3G & 0 & CNN + Trans.  \\
    BossNAS-T1 & 82.2\% & 95.8\% & 38M & $224^2$ & 8.0G & 8.0G & 8.0G & 0 & CNN + Trans. \\
    \hline
    \textbf{ShiftAddNAS-T1} & \textbf{82.7\%} & \textbf{96.1\%} & 30M & $224^2$ & 6.4G & 5.4G & 6.4G & 1.0G & Hybrid \\
    \textbf{ShiftAddNAS-T1$\boldsymbol{\uparrow}$} & \textbf{83.0\%} & \textbf{96.4\%} & 30M & $256^2$ & 8.5G & 7.1G & 8.5G & 1.4G & Hybrid  \\
    \Xhline{3\arrayrulewidth}
    \end{tabular}%
    }
  \label{tab:comp_CV}%
  \vspace{-0.7em}
\end{table*}%

\vspace{-0.5em}
\subsection{ShiftAddNAS vs. SOTA Methods on NLP Tasks}
\vspace{-0.3em}

We compare ShiftAddNAS with SOTA language models on two NLP tasks to evaluate its efficacy.
Fig. \ref{fig:comp_NLP} shows that ShiftAddNAS consistently outperforms all the baselines in terms of BLEU scores and 
FLOPs.
Specifically, ShiftAddNAS with full precision achieves \textbf{11.8\% $\sim$ 73.6\%} FLOPs reductions while offering a comparable or better BLEU score (-0.3 $\sim$ +1.1), over all the full precision baselines. 
To benchmark with Lite Transformer (8-bit) which is dedicated for mobile devices, we refer to a SOTA quantization technique \citep{banner2018scalable} for quantizing ShiftAddNAS 
to 8-bit fixed point: 
ShiftAddNAS (8-bit) achieves \textbf{+1.8 $\sim$ +4.9} BLEU scores improvements over Lite Transformer (8-bit), while offering 5.0\% $\sim$ 82.7\% FLOPs reductions, and aggressively reduces \textbf{91.6\% $\sim$ 98.4\%} FLOPs as compared to all the full-precision baselines with comparable BLEU scores (-0.1 $\sim$ +0.3).
Note that for quantized models, we follow \citep{zhou2016dorefa} to use $\text{FLOPs} \times (\text{Bit}/32)^2$ for calculating the effective FLOPs which is proportional to the number of bit-operations.
We further compare various aspects of ShiftAddNAS with all baselines in Tab. \ref{tab:comp_NLP}. As illustrated in this Table,  
ShiftAddNAS consistently outperforms the baselines, e.g., achieves up to \textbf{+2} BLEU scores improvement when comparing ShiftAddNAS (8-bit) with Lite Transformer on WMT'14 En-Fr and \textbf{69.1\%} and \textbf{69.2\%} energy and latency savings when comparing ShiftAddNAS with Transformer on WMT'14 En-De, with a comparable or even fewer model parameters and FLOPs.

\begin{figure}
\centering
\resizebox{0.9\linewidth}{!}{
    \includegraphics[width=0.9\linewidth,height=0.4\linewidth]{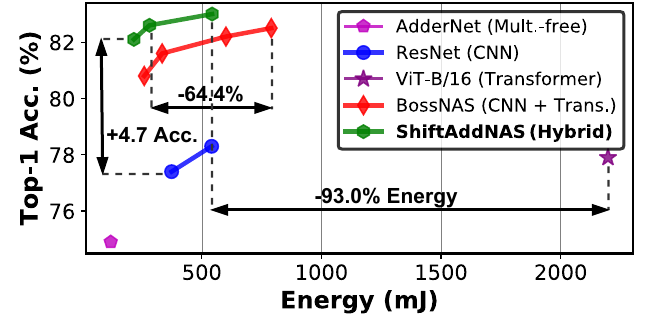}
}
\vspace{-0.9em}
\caption{Accuracy vs. energy costs of ShiftAddNAS over baselines when tested on ImageNet.}
\vspace{-2em}
\label{fig:comp_CV}
\end{figure}

\begin{table*}[t]
    \centering
    \caption{Comparison with FBNet on CIFAR-10/100 dataset.}
    \renewcommand{\arraystretch}{1.1}
    \resizebox{\textwidth}{!}{
        \begin{tabular}{cl||cc||ccc||c}
        \Xhline{3\arrayrulewidth}
        \textbf{Dataset} & \textbf{Methods} & \textbf{Accuracy} & \textbf{MACs} & \textbf{\#Mult.} & \textbf{\#Add} & \textbf{\#Shift} & \textbf{Latency Savings} \\
        \hline
        \multirow{2}[2]{*}{CIFAR-10} & FBNet & 95.09\% & 47M & 47M & 47M & 0 & - \\
          & ShiftAddNAS & 95.83\% \textbf{(+0.74\%)} & 47M & 17M & 58M & 19M & \textbf{33.80\%} \\
        \hline
        \multirow{2}[2]{*}{CIFAR100} & FBNet & 77.86\% & 55M & 55M & 55M & 0 & - \\
          & ShiftAddNAS & 78.64\% \textbf{(+0.58\%)} & 52M & 22M & 62M & 21M & \textbf{38.60\%} \\
        \Xhline{3\arrayrulewidth}
        \end{tabular}%
    }
    \label{tab:comp_FBNet}
    \vspace{-0.5em}
\end{table*}

\vspace{-0.5em}
\subsection{ShiftAddNAS vs. SOTA Methods on CV Tasks}
\vspace{-0.3em}

We further compare ShiftAddNAS over SOTA baselines on ImageNet to evaluate its effectiveness on the image classification task.
As shown in Tab. \ref{tab:comp_CV}, \ref{tab:comp_CV_complete}, ShiftAddNAS outperforms a wide range of  baselines.
Here we refer MACs as Multiply–accumulate or Shift-accumulate operations.
For example, ShiftAddNet-T0 (searched with a 4.5G MACs constraint) with 3.7G MACs achieves an improved top-1 accuracy of (1) \textbf{+5.3\% $\sim$ +26.3\%} over SOTA multiplication-free CNNs, (2) \textbf{+0.7\% $\sim$ +6.0\%} over SOTA CNNs, (3) \textbf{+0.4\% $\sim$ +7.6\%} over SOTA Transformers, (4) \textbf{+1.3\% $\sim$ +4.8\%} over SOTA CNN-Transformers, and (5) \textbf{+1.3\%, +4.7\%, and +0.4\%} over previous SOTA NAS baselines BossNAS, VITAS, and Autoformer, respectively, under a comparable or even less MACs.
Moreover, considering looser MACs constraints, we follow BossNAS to remove the downsampling in the last stage, resulting in ShiftAddNAS-T1 with a accuracy of \textbf{82.7\%} and 6.4G MACs that surpasses T2T-ViT and BoTNet-S1-59 by \textbf{+0.8\%} and \textbf{+1.0\%} at even less MACs.
By directly testing on larger input resolutions without finetuning, ShiftAddNAS-T1$\boldsymbol{\uparrow}$ (w/ $256^2$ input resolution) offers an accuracy of \textbf{83.0\%}, surpassing BossNAS-T1 and Autoformer-B by \textbf{+0.8\%} and \textbf{+0.6\%} with comparable or even less MACs, respectively.
Finally, we compare ShiftAddNAS with representative baselines of various model types in terms of accuracy and energy cost in Fig. \ref{fig:comp_CV}, where each line represents 
searched/designed models with various FLOPs constraints.
We can see that ShiftAddNAS consistently outperforms all the baselines, on average offering a \textbf{+0.8\% $\sim$ +7.7\%} higher accuracy and \textbf{24\% $\sim$ 93\%} energy savings. 
Specifically, our ShiftAddNAS on average achieves a \textbf{+0.8\%} higher accuracy and \textbf{30\%} energy savings against the most competitive NAS baseline BossNAS.
{
The searched architecture visualization is supplied to Appendix \ref{sec:arch_visualize}.
}

\begin{table}[t]
  \centering
  \caption{Ablation study of ShiftAddNAS w/ (1) naive and (2) heterogeneous weight sharing.}
  \renewcommand{\arraystretch}{1.1}
  \resizebox{\linewidth}{!}{
    \begin{tabular}{l||ccc}
    \Xhline{3\arrayrulewidth}
    \textbf{ShiftAddNAS}  & \textbf{Kendall $\tau$} & \textbf{Pearson $R$} & \textbf{Spearman $\rho$}  \\
    \hline
    w/ Naive WS  & 0.49  & 0.67  & 0.69  \\
    \textbf{w/ HWS} & \textbf{0.54}  & \textbf{0.75}  & \textbf{0.74}  \\
    \hline
    \textbf{ShiftAddNAS}  & \textbf{Top-1 Acc.} & \textbf{Energy} &
    \textbf{Latency} \\
    \hline
    w/ Naive WS  & 81.3\%  & 440mJ  &  387ms \\
    \textbf{w/ HWS} & \textbf{82.7\%}  & \textbf{413mJ}  &  \textbf{252ms} \\
    \Xhline{3\arrayrulewidth}
    \end{tabular}%
  }
  \vspace{-1em}
  \label{tab:ablation}%
\end{table}%

\subsection{\hr{ShiftAddNAS vs. SOTA under Small MACs}}
\label{sec:comp_small_MACs}

\begin{table*}[!t]
    \centering
    \caption{Comparisons of different search space variants.}
    \resizebox{0.7\linewidth}{!}{
    \begin{tabular}{l|c|c|c|c||c|c|c|c}
        \Xhline{3\arrayrulewidth}
        \multirow{2}[4]{*}{\textbf{Operators}} & \multicolumn{4}{c||}{\textbf{Our Space on ImageNet}} & \multicolumn{4}{c}{\textbf{FBNet Space on CIFAR100}}  \\
        \cline{2-9}  & \textbf{Acc.} & \textbf{Mult.} & \textbf{Add} & \textbf{Shift} & \textbf{Acc.} & \textbf{Mult.} & \textbf{Add} & \textbf{Shift}  \\
        \hline
        Attn\&Conv &  81.6\% & 5.8G & 5.8G  & 0  & 77.9\% & 55M & 55M & 0 \\
        \hline
        Shift\&Add & 76.8\% & 0.1G & 5.3G & 3.2G & 71.0\% & 3M & 48M & 35M \\
        \hline
        Attn\&Conv\&Add & 82.4\% & 5.6G & 7.2G & 0 & 78.3\% & 30M & 60M & 0  \\
        \hline
        \textbf{All} & 82.6\% & 3.6G & 4.9G  & 1.4G & 78.6\% & 22M & 62M & 21M \\
        \Xhline{3\arrayrulewidth}
    \end{tabular}%
    }
    \label{tab:control_exps}
     \vspace{-1em}
\end{table*}

\begin{figure*}[t]
    \centering
    \includegraphics[width=0.8\linewidth]{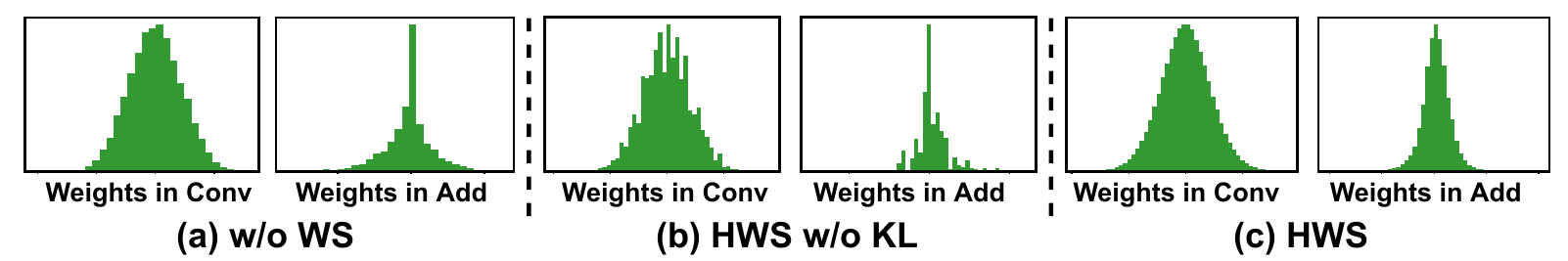}
    \vspace{-.5em}
    \caption{The weight distribution of Conv and Add layers.}
    \label{fig:distribution}
    \vspace{-1em}
\end{figure*}

To fairly compare with baselines with smaller MACs, we implemented our proposed hybrid search space and heterogeneous weight sharing (HWS) techniques on top of the FBNet search space, and evaluated the performance on CIFAR-10/100. As shown in the Tab. \ref{tab:comp_FBNet}, our ShiftAddNAS consistently boosts a +0.74\% and +0.58\% accuracy over FBNet and leads to 33.8\% and 38.6\% latency savings on CIFAR-10 and CIFAR-100, respectively.

\vspace{-0.6em}
\subsection{Ablation Studies of Heterogeneous Weight Sharing}
\vspace{-0.3em}

We conduct ablation studies on ShiftAddNAS's heterogeneous weight sharing (HWS) strategy, as shown in Tab. \ref{tab:ablation}.
\underline{\textit{First}}, for searching on ImageNet,
we use three ranking correlation metrics: Kendall $\tau$, Pearson $R$, and Spearman $\rho$, to measure the ranking correlation between ShiftAddNAS w/ and w/o HWS and find that the former leads to a higher ranking correlation than the naive WS.
\underline{\textit{Second}},
the proposed HWS leads to more accurate searched subnets. Specifically, the searched subnet achieves a \textbf{+1.4\%} higher accuracy than that of naive WS, at comparable or even smaller energy and latency costs.
Also, HWS effectively reduces the supernet size from 615M (w/o WS) to 364M (41\% savings).
This set of ablation studies validate the effectiveness of our proposed HWS strategy.
{In addition, the search cost analysis of ShiftAddNAS is placed in Appendix \ref{sec:search_cost}.}

\subsection{\hr{Ablation Studies of {\tt \textbf{Conv/Add}} Distribution}}

We visualize the weight distributions of {\tt \textbf{Conv/Add}} layers in Supernets under three scenarios, (1) w/o WS; (2) HWS w/o KL loss; (3) HWS, as shown in Fig. \ref{fig:distribution}.
We consistently observe that weights of {\tt \textbf{Conv/Add}} layers follow Gaussian and Laplacian distribution, especially when applying the introduced KL loss.

\subsection{\hr{Ablation Studies of Search Space}}

To validate the necessity of considered searchable blocks, we consider three scenarios to breakdown the search space benefits, i.e., only using:
(1) Attn \& Conv (i.e., BossNAS); 
(2) Shift \& Add;
and (3) Attn \& Conv \& Add, respectively.
Additionally, we also conduct such controlled experiments on top of FBNet search space (ignore Attn). In Tab. \ref{tab:control_exps}, we consistently see that our search space consisting of all operators outperforms (1) in terms of both accuracy and efficiency; achieves much higher accuracy than (2); and gains slightly better accuracy-efficiency trade-offs than (3).

\vspace{-0.7em}
\section{Conclusion}
\vspace{-0.5em}

We propose ShiftAddNAS for searching for multiplication-reduced NNs incorporating both powerful yet costly multiplications and efficient yet less powerful shift and add operators for marrying the best of both worlds.
ShiftAddNAS is made possible by integrating: 
(1) the first hybrid search space that incorporates both multiplication-based and multiplication-free operators;
and
(2) a novel heterogeneous weight sharing strategy that allows different operators to follow heterogeneous distributions for alleviating the dilemma of either inefficient search or inconsistent architecture ranking when searching hybrid NNs.
Extensive experiments on both NLP and CV tasks demonstrate the superior accuracy and efficiency of ShiftAddNAS's searched NNs over various SOTA baselines, opening up a new perspective in searching for more accurate and efficient NNs.

\section*{\hr{Acknowledgements}}
We would like to acknowledge the funding support from NSF SCH program (Award number: 1838873) and NIH (Award number: R01HL144683) for this project.

\nocite{langley00}

\bibliography{icml2022}
\bibliographystyle{icml2022}

\newpage
\appendix
\onecolumn


\section{Evaluate the Search Cost}
\label{sec:search_cost}

We further supply the total search cost of ShiftAddNAS on both NLP tasks and CV tasks in Table \ref{tab:NLP_cost} and \ref{tab:CV_cost}, respectively.
For NLP tasks, with one Nvidia V100 GPU, ShiftAddNAS uses on average 9.3 GPU days (Gds) for searching which is comparable to HAT \citep{wang2020hat} and 9,821$\times$ less than the Evolved Transformer \citep{so2019evolved}. For CV tasks, ShiftAddNAS uses on average 8.9 Gds for searching which is 11\% and 82\% less than DARTS \citep{liu2018darts} and BossNAS \citep{li2021bossnas}, respectively. 
In addition, we provide a concrete breakdown analysis of ShiftAddNAS search cost in Table \ref{tab:breakdown}. For NLP tasks, ShiftAddNAS needs on average 8.5 Gds for supernet training and 0.8 Gds for architecture searching; For CV tasks, ShiftAddNAS takes on average 7.7 Gds for supernet training and 1.2 Gds for architecture searching.

\begin{table}[!htbp]
    \centering
    
    \begin{minipage}{0.32\linewidth}
      \centering
      \caption{Search cost on NLP tasks.}
      \resizebox{\textwidth}{!}{
        \begin{tabular}{l|c}
            \hline
            \textbf{Methods} & \textbf{Search Cost} \\
            \hline
            Evolved Trans. & 91,334 Gds \\
            HAT & 9.3 Gds \\
            \textbf{ShiftAddNAS} & 9.3 Gds \\
            \hline
        \end{tabular}%
        \label{tab:NLP_cost}
      }
    \end{minipage}
    \begin{minipage}{0.32\linewidth}
      \centering
      \caption{Search cost on CV tasks.}
      \resizebox{\textwidth}{!}{
        \begin{tabular}{l|c}
            \hline
            \textbf{Methods} & \textbf{Search Cost} \\
            \hline
            DARTS & 50 Gds \\
            BossNAS & 10 Gds \\
            \textbf{ShiftAddNAS} & 8.9 Gds \\
            \hline
        \end{tabular}%
        \label{tab:CV_cost}
      }
    \end{minipage}
    \begin{minipage}{0.33\linewidth}
      \centering
      \caption{Breakdown analysis of the search cost of ShiftAddNAS.}
      \resizebox{\textwidth}{!}{
        \begin{tabular}{l|cc}
            \hline
              & \textbf{NLP} & \textbf{CV} \\
            \hline
            Supernet Train & 8.5 Gds & 7.7 Gds \\
            Arch. Search & 0.8 Gds & 1.2 Gds \\
            \hline
        \end{tabular}%
        \label{tab:breakdown}
      }
    \end{minipage}
\end{table}

\section{Visualization of the Heterogeneous Weight Distributions}
\label{sec:dist_visualize}

\begin{figure}[!htbp]
    \centering
    \includegraphics[width=0.98\linewidth]{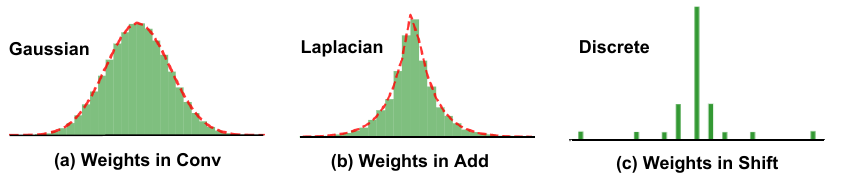}
    \caption{Visualization of the heterogeneous weight distributions in {\tt\textbf{Conv/Add/Shift}} layers.}
    \label{fig:vis_weight_dists}
\end{figure}

For better understanding of the proposed heterogeneous weight sharing strategy, we further supply the visualization of the heterogeneous weight distributions in {\tt\textbf{Conv/Add/Shift}} layers, respectively, as shown in the Fig. \ref{fig:vis_weight_dists}.

\vspace{-1em}




\section{Complete Comparison between ShiftAddNAS and SOTA on CV Tasks}

We further compare ShiftAddNAS over SOTA baselines on ImageNet to evaluate its effectiveness on the image classification task, and supply the complete comparing results to Tab. \ref{tab:comp_CV_complete} below.
Note that we report both the rank \#1 searched architecture with the highest accuracy that contains {\tt \textbf{Shift}} and {\tt \textbf{Conv}} blocks, and rank \#2 searched architecture contains additional two {\tt \textbf{Add}} blocks and achieves a 82.8\% top-1 accuracy on ImageNet with 8.4G MACs (\#Mult: 7.4G; \#Add: 9.1G; \#Shift: 0.5G).

\begin{table*}[t]
  \centering
  \caption{Comparison with SOTA baselines on ImageNet classification task.}
  \renewcommand{\arraystretch}{1.1}
  \resizebox{0.98\textwidth}{!}{
    \begin{tabular}{lcc||ccc||ccc||c}
    \Xhline{3\arrayrulewidth}
    \textbf{Model} & \textbf{Top-1 Acc.} & \textbf{Top-5 Acc.} & \textbf{Params} & \textbf{Res.}  & \textbf{MACs} & \textbf{\#Mult.} & \textbf{\#Add} & \textbf{\#Shift} & \textbf{Model Type} \\
    \Xhline{3\arrayrulewidth}
    BNN & 55.8\% & 78.4\% & 26M & $224^2$ & 3.9G & 0.1G & 3.9G & 3.8G & Mult.-free  \\
    AdderNet & 74.9\% & 91.7\% & 26M & $224^2$ & 3.9G & 0.1G & 7.6G & 0 & Mult.-free \\
    AdderNet-PKKD & 76.8\% & 93.3\% & 26M & $224^2$ & 3.9G & 0.1G & 7.6G & 0 & Mult.-free \\
    DeepShift-Q & 70.7\% & 90.2\% & 26M & $224^2$ & 3.9G & 0.1G & 3.9G & 3.8G & Mult.-free \\
    DeepShift-PS & 71.9\% & 90.2\% & 52M & $224^2$ & 3.9G & 0.1G & 3.9G & 3.8G & Mult.-free  \\
    ShiftAddNet & 72.3\% & - & 64M & $224^2$ & 10G & 0.1G & 16G & 3.9G & Mult.-free  \\
    \hline
    ResNet-50 & 76.1\% & 92.9\% & 26M & $224^2$ & 3.9G & 3.9G & 3.9G & 0 & CNN  \\
    ResNet-101 & 77.4\% & 94.2\% & 45M & $224^2$ & 7.6G & 7.6G & 7.6G & 0 & CNN \\
    SENet-50 & 79.4\% & 94.6\% & 26M & $224^2$ & 3.9G & 3.9G & 3.9G & 0 & CNN \\
    SENet-101 & 81.4\% & 95.7\% & 45M & $224^2$ & 7.6G & 7.6G & 7.6G & 0 & CNN \\
    \hline
    ViT-B/16 & 77.9\% & - & 86M & $384^2$ & 18G & 18G & 17G & 0 & Transformer  \\
    ViT-L/16 & 76.5\% & - & 304M & $384^2$ & 64G & 64G & 63G & 0 & Transformer \\
    DeiT-T & 74.5\% & - & 6M & $224^2$ & 1.3G & 1.3G & 1.3G & 0 & Transformer \\
    DeiT-S & 81.2\% & - & 22M & $224^2$ & 4.6G & 4.6G & 4.6G & 0 & Transformer  \\
    VITAS & 77.4\% & 93.8\% & 13M & $224^2$ & 2.7G & 2.7G & 2.7G & 0 & Transformer  \\
    Autoformer-S & 81.7\% & 95.7\% & 23M & $224^2$ & 5.1G & 5.1G & 5.1G & 0 & Transformer  \\
    \hline
    BoT-50 & 78.3\% & 94.2\% & 21M & $224^2$ & 4.0G & 4.0G & 4.0G & 0 & CNN + Trans.  \\
    BoT-50 + SE & 79.6\% & 94.6\% & 21M & $224^2$ & 4.0G & 4.0G & 4.0G & 0 & CNN + Trans. \\
    HR-NAS & 77.3\% & - & 6.4M & $224^2$ & 0.4G & 0.4G & 0.4G & 0 & CNN + Trans. \\
    BossNAS-T0 & 80.5\% & 95.0\% & 38M & $224^2$ & 3.5G & 3.5G & 3.5G & 0 & CNN + Trans. \\
    BossNAS-T0 + SE & 80.8\% & 95.2\% & 38M & $224^2$ & 3.5G & 3.5G & 3.5G & 0 & CNN + Trans.  \\
    \hline
    \textbf{ShiftAddNAS-T0} & \textbf{82.1\%} & \textbf{95.8\%} & 30M & $224^2$ & 3.7G & 2.7G & 3.8G & 1.0G & Hybrid  \\
    \textbf{ShiftAddNAS-T0$\boldsymbol{\uparrow}$} & \textbf{82.6\%} & \textbf{96.2\%} & 30M & $256^2$ & 4.9G & 3.6G & 4.9G & 1.4G & Hybrid  \\
    \Xhline{3\arrayrulewidth}
    T2T-ViT-19 & 81.9\% & - & 39M & $224^2$ & 8.9G & 8.9G & 8.9G & 0 & Transformer  \\
    TNT-S & 81.3\% & 95.6\% & 24M & $224^2$ & 5.2G & 5.2G & 5.2G & 0 & Transformer \\
    Autoformer-B & 82.4\% & 95.7\% & 54M & $224^2$ & 11G & 11G & 11G & 0 & Transformer  \\
    \hline
    BoTNet-S1-59 & 81.7\% & 95.8\% & 28M & $224^2$ & 7.3G & 7.3G & 7.3G & 0 & CNN + Trans.  \\
    BossNAS-T1 & 82.2\% & 95.8\% & 38M & $224^2$ & 8.0G & 8.0G & 8.0G & 0 & CNN + Trans. \\
    \hline
    \textbf{ShiftAddNAS-T1} & \textbf{82.7\%} & \textbf{96.1\%} & 30M & $224^2$ & 6.4G & 5.4G & 6.4G & 1.0G & Hybrid \\
    \textbf{ShiftAddNAS-T1$\boldsymbol{\uparrow}^{1}$} & \textbf{83.0\%} & \textbf{96.4\%} & 30M & $256^2$ & 8.5G & 7.1G & 8.5G & 1.4G & Hybrid  \\
    \textbf{ShiftAddNAS-T1$\boldsymbol{\uparrow}^{2}$} & \textbf{82.8\%} & \textbf{96.2\%} & 30M & $256^2$ & 8.4G & 7.4G & 9.1G & 0.5G & Hybrid \\
    \Xhline{3\arrayrulewidth}
    \end{tabular}%
    }
  \label{tab:comp_CV_complete}%
  \vspace{-1.em}
\end{table*}%

\section{Visualization of the Searched Architecture}
\label{sec:arch_visualize}

\begin{figure}[t]
    \centering
    \includegraphics[width=\linewidth]{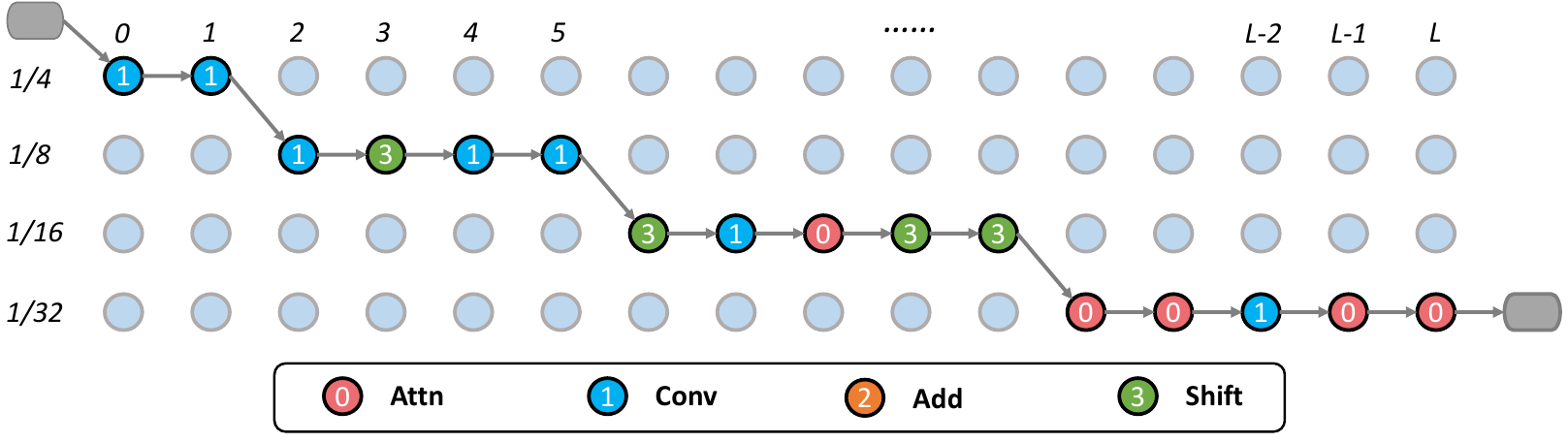}
    \caption{Visualization of the searched architecture ShiftAddNAS-T1$\boldsymbol{\uparrow}^{1}$ with 83\% top-1 test accuracy on ImageNet.}
    \label{fig:vis_arch_1}
\end{figure}

\begin{figure}[t]
    \centering
    \includegraphics[width=\linewidth]{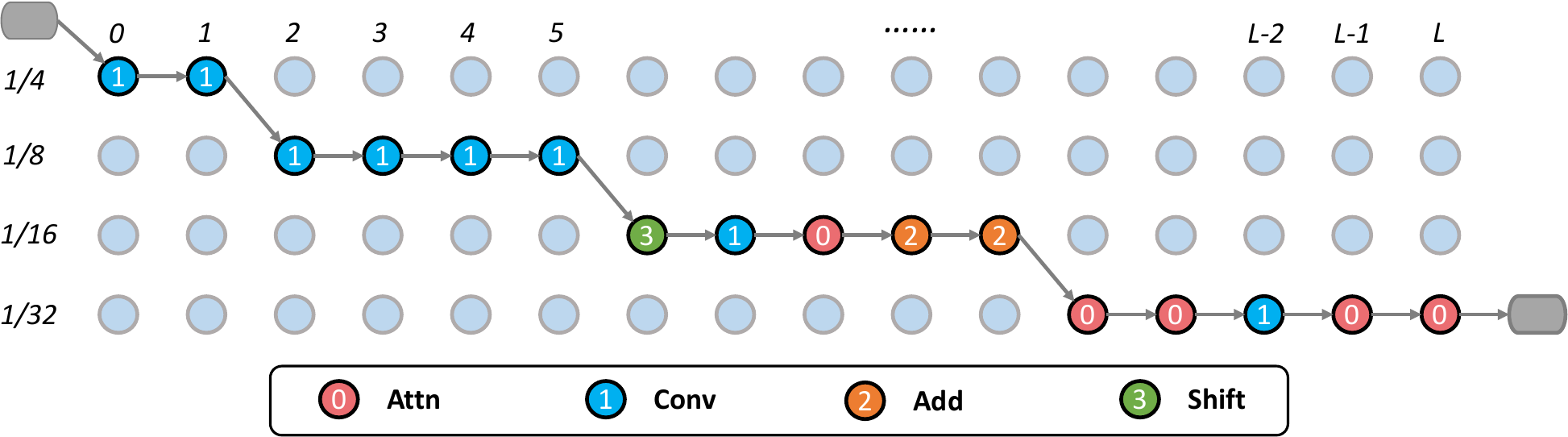}
    \caption{Visualization of the searched architecture ShiftAddNAS-T1$\boldsymbol{\uparrow}^{2}$ with 82.8\% top-1 test accuracy on ImageNet.}
    \label{fig:vis_arch_2}
\end{figure}

For better understanding of the searched architecture, we provide the visualization of the searched architecture ShiftAddNAS-T1$\boldsymbol{\uparrow}^{1}$ and ShiftAddNAS-T1$\boldsymbol{\uparrow}^{2}$ in Fig. \ref{fig:vis_arch_1} and Fig. \ref{fig:vis_arch_2}, respectively. 
The searched architecture ShiftAddNAS-T1$\boldsymbol{\uparrow}^{1}$ contains four {\tt \textbf{Shift}} blocks, and achieves 83\% top-1 test accuracy on ImageNet with 8.5G MACs (\#Mult: 7.1G; \#Add: 8.5G; \#Shift: 1.4G).
The searched architecture ShiftAddNAS-T1$\boldsymbol{\uparrow}^{2}$ contains two {\tt \textbf{Add}} blocks and one {\tt \textbf{Shift}} block, and achieves 82.8\% top-1 test accuracy on ImageNet with 8.4G MACs (\#Mult: 7.4G; \#Add: 9.1G; \#Shift: 0.5G).
Moreover, the searched architecture prefers {\tt\textbf{Conv}} as early blocks while consider {\tt\textbf{Attn}} as later blocks, which is also consistent with the previous empirical observation that early convolutions help the overall performance \citep{xiao2021early}.

\textbf{More discussion for the searched architectures.}
The performance of the searched top architectures are quite close (e.g., the accuracy difference between rank \#1 and rank \#2 is smaller than 0.2\%), and we also quantitatively measure the ratio breakdown of different operators in our searched top-10 architectures ({\tt \textbf{Attn}}: 30\%; {\tt \textbf{Conv}}: 43\%; {\tt \textbf{Shift}}: 15\%; {\tt \textbf{Add}}: 12\%). We see that the overall ratio of {\tt \textbf{Add}} operators is quite comparable with that of {\tt \textbf{Shift}} operators, thus our understanding is that the searched top architectures benefit from a relatively more balanced combination/contribution of all operators to push forward the frontier of accuracy-efficiency tradeoff. Furthermore, showing such a ratio breakdown in searched top architectures further validates the necessity of adopting a hybrid search space.
Finally, another insight from the general trends of the operator ratio breakdown is that, as the ranking of searched architectures drops, the ratio of Shift and Add operators will increase, which is consistent with our experiments, i.e., a purely ShiftNet or AdderNet will inevitably lead to accuracy drop. The above analysis also helps to validate the correctness of our search algorithm.

\section{Detailed Search and Training Settings}
\label{sec:exp_setting}

\underline{\textit{For NLP tasks,}} 
after training the supernet for 40K steps, we adopt an evolutionary algorithm \citep{wang2020hat} to search for subnets with various latency and FLOPs constraints ranging from 1.5G to 4.5G for 30 steps with a population of 125, a crossover population of 50, and a mutation population of 50 with a probability of 0.3.
During search, measuring latency for each chosen subnet can be time-consuming. Instead, we estimate the latency using a three-layer NN trained with encoding architecture parameters as features and measured latency as labels following \citep{wang2020hat}.
The latency predictor is accurate with an average prediction error of $<$ 5\%.
The searched subnets are then retrained from scratch for another 40K steps with an Adam optimizer and a cosine learning rate (LR) scheduler, where the LR is linearly warmed up from $10^{-7}$ to $10^{-3}$ and then annealed (same for training supernets).
\underline{\textit{For CV tasks,}} 
we conduct an evolutionary search with FLOPs constraints for 20 steps with a population of 50, a crossover population of 25, and a mutation population of 25 with a probability of 0.2 following \citep{AutoFormer}.
We train both the supernet and searched subnets using the same recipe and hyperparameters as DeiT \citep{DeiT}.
Note that the position encoding in the attention blocks is replaced with a depthwise convolution following \citep{li2021bossnas} for reducing the computational complexity. 



\end{document}